%% file: main.tex
\documentclass[10pt,twocolumn,letterpaper]{article}

\usepackage{cvpr}              
\input{preamble}
\definecolor{cvprblue}{rgb}{0.21,0.49,0.74}
\usepackage[pagebackref,breaklinks,colorlinks,allcolors=cvprblue]{hyperref}

\def\paperID{20113} 
\def\confName{CVPR}
\def\confYear{2026}
\newcommand{\benchmark}{\textsc{DanceTypesBenchmark}\xspace}
\newcommand{\tool}{\textsc{DanceMatch}\xspace}
\newcommand{\task}{\textsc{Dance Fingerprinting}\xspace}
\newcommand{\retrieval}{\textsc{Dance Retrieval Engine (DRE)}\xspace}
\title{Learning Quantised Structure-Preserving Motion Representations \\ for Dance Fingerprinting}

\author{Arina Kharlamova$^{*}$\\
Department of Computer Science\\
MBZUAI, Abu Dhabi, UAE\\
{\tt\small arina.kharlamova@mbzuai.ac.ae}
\and
Bowei He\thanks{Equal contribution}~~\\
Department of Computer Science\\
City University of Hong Kong, Hong Kong, China \\
{\tt\small boweihe2-c@my.cityu.edu.hk}
\and
Chen Ma\\
Department of Computer Science\\
City University of Hong Kong, Hong Kong, China \\
{\tt\small chenma@my.cityu.edu.hk}
\and
Xue Liu\\
Department of Computer Science\\
MBZUAI, Abu Dhabi, UAE\\
{\tt\small steve.liu@mbzuai.ac.aeg}
}

\begin{document}
\maketitle
\input{sec/0_abstract}    
\input{sec/1_introduction}
\input{sec/2_related_work}
\input{sec/3_background}
\input{sec/4_method}

\input{sec/5_experiments}
\input{sec/6_conclusion}
{
    \small
    \bibliographystyle{ieeenat_fullname}
    \bibliography{main}
}


\end{document}

%% file: sec/0_abstract.tex
\begin{abstract}
We present \tool, an end-to-end framework for motion-based dance retrieval, the task of identifying semantically similar choreographies directly from raw video, defined as \task. While existing motion analysis and retrieval methods can compare pose sequences, they rely on continuous embeddings that are difficult to index, interpret, or scale. In contrast, \tool constructs compact, discrete motion signatures that capture the spatio-temporal structure of dance while enabling efficient large-scale retrieval. Our system integrates Skeleton Motion Quantisation (SMQ) with Spatio-Temporal Transformers (STT) to encode human poses, extracted via Apple’s CoMotion, into a structured motion vocabulary. We further design \retrieval, which performs sub-linear retrieval using a histogram-based index followed by re-ranking for refined matching. To facilitate reproducible research, we release \benchmark, a pose-aligned dataset annotated with quantised motion tokens. Experiments demonstrate robust retrieval across diverse dance styles and strong generalisation to unseen choreographies, establishing a foundation for scalable motion fingerprinting and quantitative choreographic analysis.
\end{abstract}

%% file: sec/1_introduction.tex
\section{Introduction}
\label{sec:introduction}

Music can be identified within seconds using audio fingerprinting systems such as \emph{Shazam} \cite{keoikantse_mogonediwa_86e93822, joren_six_91435ba8, furkan_yesiler_fcdf4565, salvatore_serrano_d10952da, aar_n_l_pez_garc_a_65fa97fd, mandla_vamshikrishna_ce788cfc}; yet no equivalent end-to-end framework exists for recognising or retrieving dance performances from video.
In the contemporary media ecosystem dominated by short-form video \cite{yanjie_pan_632ac205, xuanchen_wang_ed72dae9, jiahao_wang_b55ca158, daniel_c__castro_4f2eee81} platforms \cite{jiahao_wang_b55ca158} (\eg \emph{TikTok} \cite{benjamin_steel_758dcd17}, \emph{Instagram Reels} and \emph{Youtube Shorts} \cite{wendi_chen_b92959b5}) dance sequences have become a pervasive mode of cultural expression and a rich source of motion data, as well as a primary medium of cultural production and reuse \cite{alexandra_harlig_c698345e, nhat_le_cdc8366d}.

Millions of choreographies are created, remixed, and circulated daily, yet there remains no mechanism to index or attribute motion content by its spatio-temporal pattern \cite{saad_hassan_7faef0fa, m_reshma_e5b301cb, bo_han_6bd38adc, xinjian_zhang_048f2250, chayapatr_archiwaranguprok_d8679fca, yan_wang_19daa95b, nhat_le_cdc8366d}. The absence of such capability underlies not only technical limitations in motion understanding, but also emerging disputes around ownership and authorship of choreographic material, exemplified by recent copyright litigations involving digital dance representations in commercial games \cite{gbenga_odugbemi_1afb9494, zachary_crane_a24c2eae}.
However, despite the ubiquity of such content, there remains no mechanism to query a video by its movement alone: given a dance clip, to efficiently locate semantically similar choreographies or stylistically related motions in a large corpus. These developments highlight the growing need for reliable tools to analyse, identify, and retrieve human motion from large-scale video collections.

\begin{figure}[t]
  \centering
  \includegraphics[width=\linewidth]{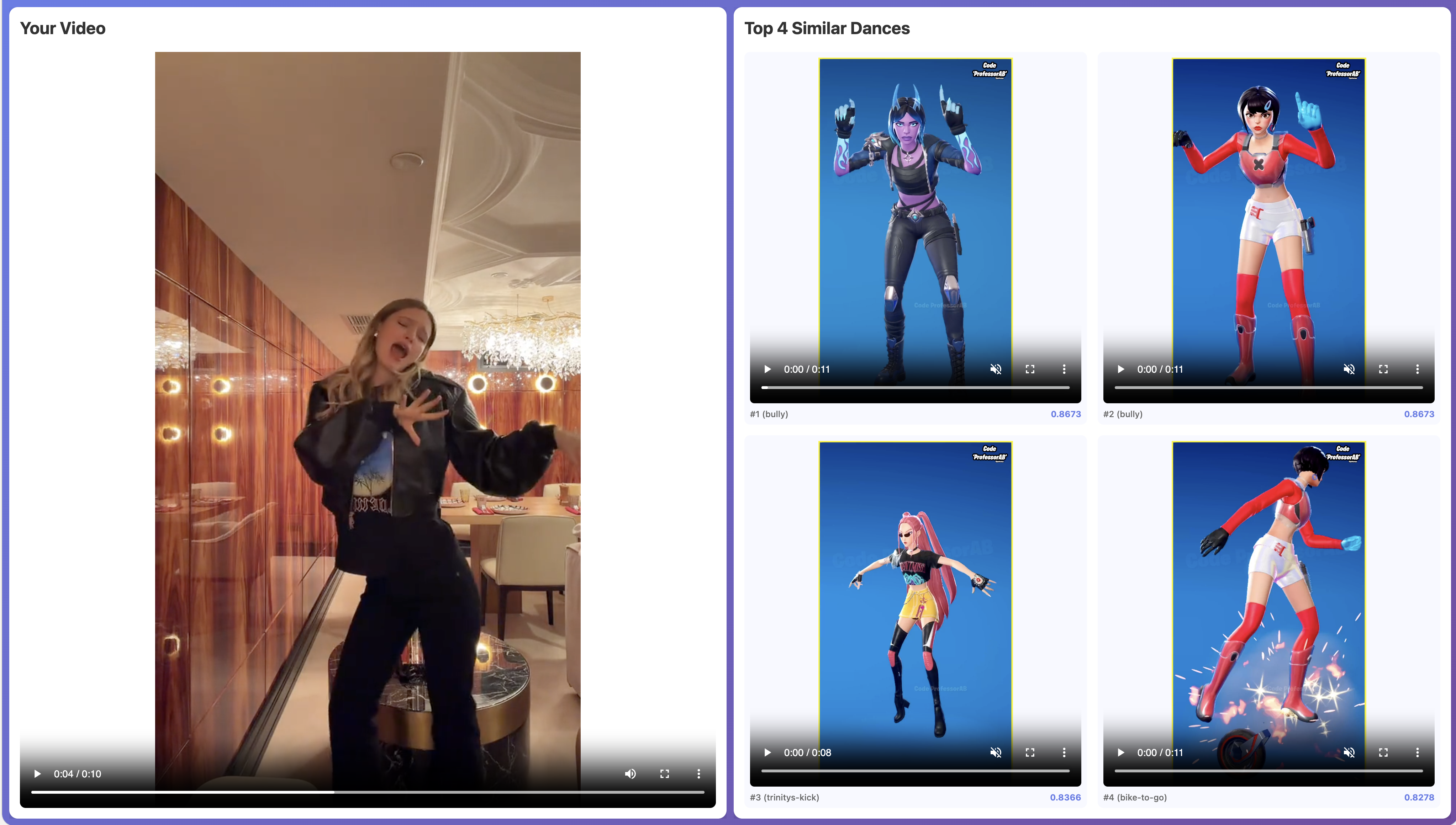}
  \caption{
        \emph{Illustration of the proposed end-to-end framework for motion-based dance retrieval.} The system processes an input dance video into a structured motion representation and retrieves top-matching dances through alignment-based similarity.
    }
  \label{fig:preview}
\end{figure}

We address this gap by introducing a novel formulation of motion-based dance retrieval, which we term \task, a motion analogue of audio fingerprinting that retrieves semantically similar dance performances directly from raw video. Unlike traditional motion analysis methods that rely on continuous embeddings or supervised labels \cite{shuhei_tsuchida_96b1e92d, bo_han_6bd38adc, k__endo_4e9c1084, kaixing_yang_9ed66893, ming_huang_d48f1e5a}, \task seeks to construct compact, discrete, and interpretable motion signatures that capture the geometric and temporal essence of a performance while enabling efficient large-scale retrieval.

Existing approaches for motion representation, including transformer-based models such as MotionBERT and Spatio-Temporal Transformers (STT) \cite{ming_huang_d48f1e5a, wentao_zhu_00b3e0bf}, achieve strong results in prediction and reconstruction tasks but are ill-suited for retrieval. Their continuous latent embeddings are high-dimensional, non-discrete, and difficult to index efficiently; moreover, they lack interpretability and offer no symbolic structure for motion comparison.
In contrast, a \task system should produce compact, invariant, and interpretable motion signatures that enable large-scale matching while preserving spatio-temporal structure.

To address this new problem, we propose the first end-to-end \tool framework for motion-based dance retrieval. As illustrated in~\cref{fig:preview}, our system converts raw dance videos into discrete motion tokens through a structure-preserving quantisation process, and retrieves similar performances using alignment-based metrics and hierarchical indexing.
We then design \retrieval, which performs efficient sub-linear retrieval using a histogram-based index, followed by subsequent re-ranking for refined matching (\S\ref{sec:retrieval}).
To support evaluation and future research, we curate \benchmark, the first dataset of pose-aligned skeleton sequences annotated with quantised motion tokens.
Together, these components establish the first reproducible framework for large-scale dance fingerprinting (\eg retrieving and analysing motion similarity directly from video in a way that parallels audio fingerprinting for music).

We summarize our contributions in this paper as follows:
\begin{itemize}
    \item We introduce the task of motion-based dance retrieval, or \task, which retrieves semantically similar dance performances directly from video (\S\ref{sec:preliminaries}).
    \item We propose \tool a structure-preserving motion quantisation framework
    (\S\ref{sec:dance_match}).
    \item We design a scalable \retrieval combining alignment-based metrics with hierarchical indexing, enabling efficient large-scale search across thousands of motion sequences (\S\ref{sec:retrieval}).
    \item We curate and release \benchmark, the first dance-retrieval dataset of pose-aligned skeleton sequences with quantised motion-token annotations, establishing a benchmark for reproducible evaluation (\S\ref{sec:benchmark}).
\end{itemize}

%% file: sec/2_related_work.tex
\section{Related Works}
\label{sec:related_work}

\subsection{RGB-to-3D Human Pose and Mesh Estimation}

Estimating articulated human motion from monocular RGB video has long been a central challenge in computer vision and human modelling. Recent learning-based pipelines have advanced this task by integrating temporal reasoning and biomechanical constraints. VIBE~\cite{kocabas2020vibevideoinferencehuman} formulates video-based inference of human pose and shape as a temporal regression problem, employing a recurrent network over SMPL~\cite{10.1145/2816795.2818013} parameters to yield smooth, coherent 3D reconstructions from unconstrained RGB input. More recently, BioPose~\cite{koleini2025bioposebiomechanicallyaccurate3dpose} introduces explicit biomechanical modelling into the estimation process, enforcing physically plausible joint limits and muscular dynamics, which significantly improves anatomical consistency and motion realism. These methods constitute the foundational layer upon which skeleton-based recognition pipelines are constructed, providing structured 3D representations that enable semantic reasoning about motion.

\subsection{Skeleton-Based Action Recognition from Monocular RGB}

Skeleton-driven action recognition methods build upon 2D or pseudo-3D joint representations extracted directly from RGB frames. The approach of Silva \emph{et al.}~\cite{s21134342} employs an image-based spatio–temporal encoding of skeletal dynamics, which can be processed by standard convolutional backbones to capture short-term dependencies. In contrast, Gupta \emph{et al.}~\cite{gupta16_efficient_retrieval} propose a retrieval-oriented framework that aligns human motion sequences under varying temporal dynamics, enabling flexible comparison of motion patterns across individuals and contexts. Such methods serve as a bridge between low-level pose estimation and higher-level activity recognition, though they remain limited by their reliance on monocular visual cues and the lack of explicit 3D temporal coherence.

\subsection{Skeleton-Based Action Recognition from 3D Joint Trajectories}

With the advent of robust 3D pose estimators, research has shifted towards exploiting explicit 3D joint trajectories for temporal action understanding. Recent transformer-based models, such as ST-RTR~\cite{mehmood2024humanactionrecognitionhar}, model spatial and temporal relationships jointly across the skeletal graph, capturing both inter-joint dependencies and longer-term dynamics. Similarly, Wang \emph{et al.}~\cite{wang2024skeletonbasedactionrecognitionspatialstructural} propose a spatial–structural graph convolutional network that leverages the intrinsic topology of the human skeleton for improved discriminative power. These architectures demonstrate the importance of structural priors in encoding motion semantics, suggesting that relational reasoning over 3D joints can outperform frame-level or 2D-based representations.

\subsection{Recognition of Dance Movements in Human Motion Analysis}

The recognition and retrieval of dance movements constitute a specialised subset of human motion analysis, distinguished by fine-grained temporal alignment and stylistic expressivity. Yan \emph{et al.}~\cite{Yan2022-fr} present a deep learning model optimised for real-time dance action recognition in streaming contexts, focusing on robustness to motion variability and low-latency inference. Tsuchida and Shuhei~\cite{Tsuchida2019-ba} introduce the \emph{Query-by-Dancing} paradigm, in which body-motion similarity serves as the retrieval key for music or choreography datasets.  Earlier work by Oveneke \emph{et al.}~\cite{Oveneke2012-xm} achieves invariance to anthropometric and temporal differences through motion normalisation, enabling cross-subject comparison of dance sequences.  Collectively, these studies highlight the challenges of capturing nuanced temporal and stylistic variation in motion, motivating the need for domain-adaptive representations that couple biomechanical accuracy with expressive fidelity.

%% file: sec/3_background.tex
\section{Preliminaries}
\label{sec:preliminaries}

The task of dance fingerprinting concerns the automatic identification and retrieval of choreographic patterns from raw video streams. Given an input sequence depicting human movement, the objective is to recover either a canonical identifier of the dance or a ranked set of semantically similar performances. At its core, the problem requires a representation that preserves choreographic structure while remaining robust to surface variation in appearance, performer morphology, and recording conditions. Unlike conventional classification, which maps motion to discrete categories, fingerprinting demands relational discrimination: the ability to measure nuanced correspondences between sequences that may differ in timing, articulation, or stylistic execution.

Thus, We consider the problem of \task: identifying or retrieving structurally similar dance performances from large-scale video corpora. Given an input video sequence depicting human motion, the system must produce either (i) a unique dance identifier corresponding to a canonical choreography, or (ii) a ranked list of semantically or kinematically similar dance segments. The central challenge is to design a representation that is invariant to superficial factors (\eg camera viewpoint, lighting, performer identity) yet sensitive to choreographic structure and temporal alignment. Formally,

\begin{equation}
    V = \{v_t\}_{t=1}^{T}
\end{equation}

denote a temporally ordered sequence of visual observations, where each frame \(v_t\) may be represented as RGB imagery, optical flow, or a 3-D skeletal pose vector. The fingerprinting model. 

\begin{equation}
    f : V \to Q
\end{equation}

maps the video \(V\) to a discrete sequence of motion tokens 

\begin{equation}
    Q = \{q_1, q_2, \ldots, q_L\}, \quad q_i \in \mathcal{A},
\end{equation}

where \(\mathcal{A}\) is a finite codebook learned or predefined. The token sequence \(Q\) serves as a compact, canonical \textit{motion signature} that abstracts the original choreography.

%% file: sec/4_method.tex
\section{Methodology}
\label{sec:method}

\begin{figure*}[t]
  \centering
  \includegraphics[width=0.9\linewidth]{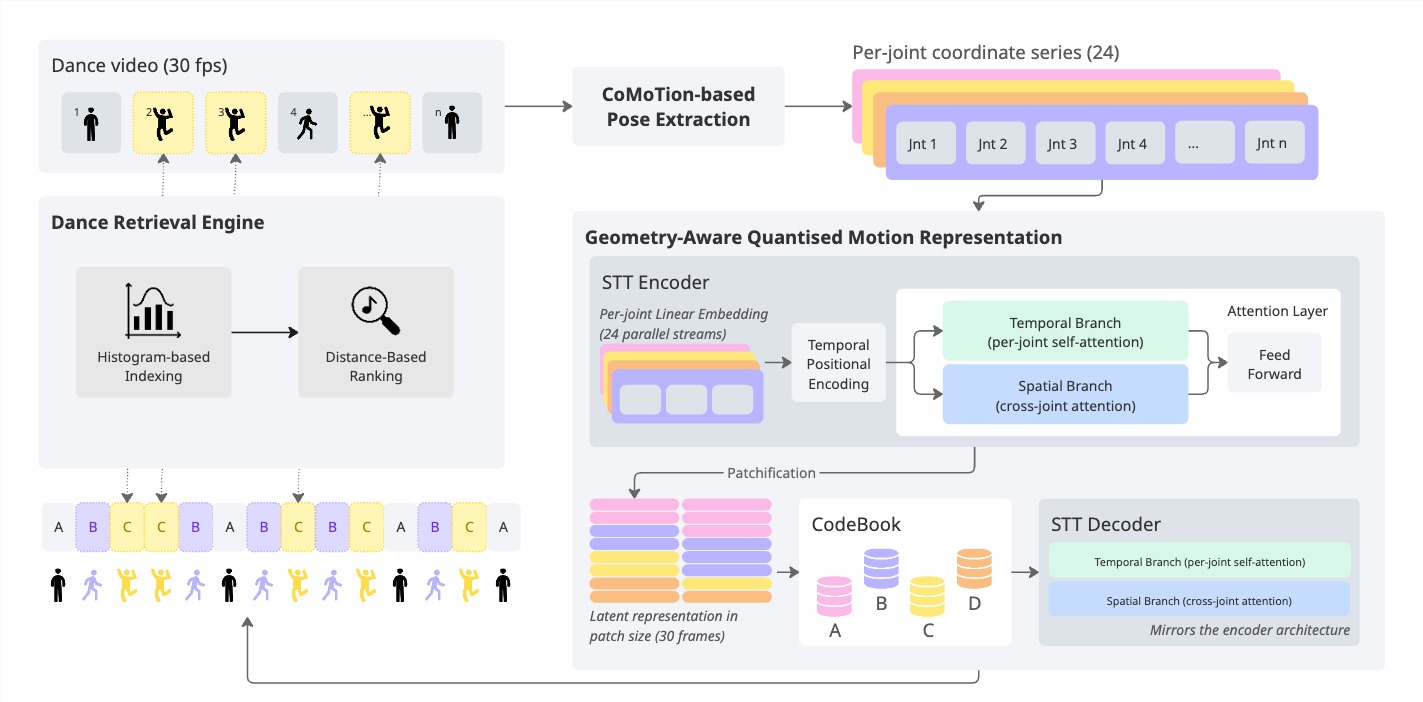}
  \caption{
  \textbf{Overview of the \tool\ framework.}
  The Spatio-Temporal Transformer (STT) encoder extracts per-joint latent embeddings
  from skeletal motion sequences.
  Latent patches are discretised via vector quantisation into motion tokens using a learnable codebook.
  The STT decoder reconstructs the motion sequence from the quantised representation.
  Immediate dead-code revival and EMA-based updates ensure stable codebook utilisation and prevent mode collapse.
  }
  \label{fig:architecture}
\end{figure*}

As illustrated in~\cref{fig:architecture}, the \tool\ framework learns a discrete representation of human motion from dance videos and then proceeds with the retrieval step. 
Two stages, \emph{Motion Quantisation via \tool} and \emph{Scalable \retrieval}, constitute the methodological core of our framework, jointly realising a unified representation–retrieval pipeline for structured human motion. 
Building upon the foundations of \emph{Skeleton Motion Quantisation} (SMQ)~\cite{gökay2025skeletonmotionwordsunsupervised} and \emph{Spatio–Temporal Transformers} (STT)~\cite{aksan2021spatiotemporaltransformer3dhuman}, we introduce an integrated formulation that enables stable large-scale optimisation and interpretable discrete motion encoding. 
The learned \emph{codebook} discretises motion representations into compact motion tokens that capture both spatial and temporal dependencies, facilitating scalable retrieval and controllable synthesis. 
Our design further re-architects the SMQ–STT~\cite{gökay2025skeletonmotionwordsunsupervised,aksan2021spatiotemporaltransformer3dhuman} interaction through codebook-aware attention, balanced commitment dynamics, and progressive quantisation warm-up (\eg mechanisms that collectively sustain high codebook utilisation and prevent mode collapse under long-sequence training). 
Together, these refinements expand the representational and temporal capacity of vector-quantised motion models, allowing stable optimisation over substantially longer and denser motion sequences while preserving reconstruction fidelity and token diversity as consistently observed in our evaluations (Sec.~\ref{sec:experiments}).

\subsection{Comotion based pose extraction}
The system begins by extracting 3D skeletal trajectories using the CoMotion-based pose estimator~\cite{newell2025comotion}, which suits our setting as it performs reliable multi-person tracking from monocular RGB videos without requiring additional calibration or model adaptation. Moreover, CoMotion offers broad hardware compatibility (\eg running efficiently on both CUDA-enabled GPUs and CoreML devices \cite{githubGitHubTucan9389PoseEstimationCoreML, appleDetecting}). This preprocessing stage is used as-is, yet can be replaced by any equivalent skeletal estimator \cite{wu2021graph, guo2020pinetposeinteractingnetwork, choi2025cooperativeinferencerealtime3d, s25082409, Cheng_2023, li2020hmorhierarchicalmultipersonordinal, mehta2018singleshotmultiperson3dpose} if desired.

\subsection{Motion Quantisation via \tool}
\label{sec:dance_match}
\tool\ performs \emph{motion quantisation} on human skeleton sequences by coupling a Spatio-Temporal Transformer (STT) \cite{aksan2021spatiotemporaltransformer3dhuman} with a vector-quantisation (VQ) bottleneck. The system learns a discrete motion vocabulary that compresses continuous skeletal trajectories into a finite set of interpretable motion tokens while preserving both spatial structure and temporal coherence.

Given an input sequence of 3D joint coordinates 
$X \in \mathbb{R}^{N \times C \times T \times V}$,
where $N$ is the batch size, $C$ the coordinate dimension, $T$ the number of frames, and $V$ the number of joints, 
\tool\ encodes the sequence using an STT encoder, discretises the latent representation via a learnable codebook, and reconstructs it using an STT decoder:
\begin{equation}
    \hat{X} = \text{STTDecoder}\big(\text{Quantize}(\text{STTEncoder}(X))\big).
\end{equation}
The encoder and decoder employ parallel dual-branch attention to model spatial and temporal dependencies simultaneously, ensuring that cross-joint and intra-joint dynamics are captured without sequential interference.

\textbf{Spatio-Temporal Transformer Encoding.}
Each joint $v$ is first projected independently into a latent space through a joint-specific linear layer, enabling joint-wise feature disentanglement. Temporal self-attention operates over the motion trajectory of each joint, capturing dynamic evolution, while spatial attention models inter-joint relationships across the skeleton. 
Both branches receive the original feature input rather than sequential outputs, preventing information bottlenecks. 
The combined representation is refined through per-joint feed-forward networks and normalisation layers, yielding latent codes 
$Z \in \mathbb{R}^{N \times V \times D \times T}$.

\textbf{Quantisation of Motion Patches.}
To obtain discrete representations, the latent tensor $Z$ is divided into temporal patches of length $P$ frames, reshaped to 
$p_i \in \mathbb{R}^{P \times (V \cdot D)}$. 
Each patch is assigned to its nearest codebook entry 
$c_k \in \mathcal{C} = \{c_k\}_{k=1}^{K}$ via Euclidean distance:
\begin{equation}
    k_i = \arg\min_{k} \|p_i - c_k\|_2, \quad q_i = c_{k_i}.
\end{equation}
Quantisation is made differentiable through a straight-through estimator:
\begin{equation}
    \tilde{q}_i = p_i + \text{sg}[q_i - p_i],
\end{equation}
where $\text{sg}[\cdot]$ denotes the stop-gradient operator. 
The quantised sequence $\tilde{Q}$ is passed to the STT decoder, which reconstructs the motion trajectory in the original space.

\textbf{Codebook Optimisation.}
The codebook is updated using Exponential Moving Averages (EMA) of code assignments:
\begin{equation}
    \begin{aligned}
    n_k^{(t)} = \alpha n_k^{(t-1)} + (1 - \alpha)\!\!\sum_i \mathbb{1}[k_i=k], \\ 
    m_k^{(t)} = \alpha m_k^{(t-1)} + (1 - \alpha)\!\!\sum_{i:k_i=k}\!p_i,
    \end{aligned}
\end{equation}
\begin{equation}
    c_k^{(t)} = \frac{m_k^{(t)}}{\max(n_k^{(t)}, \epsilon)}.
\end{equation}
To prevent code underutilisation, any code unused within an epoch is replaced by a randomly selected latent patch from the current batch. 
This \emph{immediate revival} strategy maintains high codebook utilisation (\textgreater80\%) and stabilises training by mitigating mode collapse.

\textbf{Codebook Utilisation.}
We monitor codebook health via the \emph{usage ratio}, defined as
\begin{equation}
\text{Usage} = \frac{|{k : n_k > 0}|}{K} \times 100
\end{equation}
A well-trained codebook typically exhibits utilisation above 80\%, indicating effective coverage of the latent space. Maintaining high usage prevents representational collapse and ensures diverse code participation throughout training.

\textbf{Loss Functions.}
The training objective combines structural reconstruction, commitment, and entropy regularisation terms:
\begin{equation}
    \mathcal{L}_{\text{total}}
    = \lambda \mathcal{L}_{\text{rec}}
    + \mathcal{L}_{\text{commit}}
    + \beta \mathcal{L}_{\text{codebook}}
    - \gamma \mathcal{L}_{\text{entropy}}.
\end{equation}
The reconstruction loss $\mathcal{L}_{\text{rec}}$ measures inter-joint distance error, enforcing pose-structure consistency invariant to global rotation and translation. 
The commitment loss encourages latent vectors to remain close to their assigned codebook entries, while entropy regularisation promotes uniform usage across the motion vocabulary. 
Empirically, a higher reconstruction weight ($\lambda=1.0$) yields stable convergence for large-scale datasets, differing from prior works that use smaller weights due to limited data diversity.

\textbf{Training Protocol.}
\tool\ adopts a warm-up stage in which codebook updates are temporarily suspended, allowing the encoder–decoder pair to form meaningful latent structures before quantisation. 
Subsequent training employs the Adam optimiser with learning rate $10^{-4}$, gradient clipping at $\|\nabla\|_{\max}=1.0$, batch size 16 per GPU, and EMA decay $\alpha=0.5$. 
These settings achieve balanced codebook utilisation and robust convergence across extended motion sequences.

\subsection{Scalable \retrieval}
\label{sec:retrieval}

\begin{figure}[t]
\centering
\includegraphics[width=\linewidth]{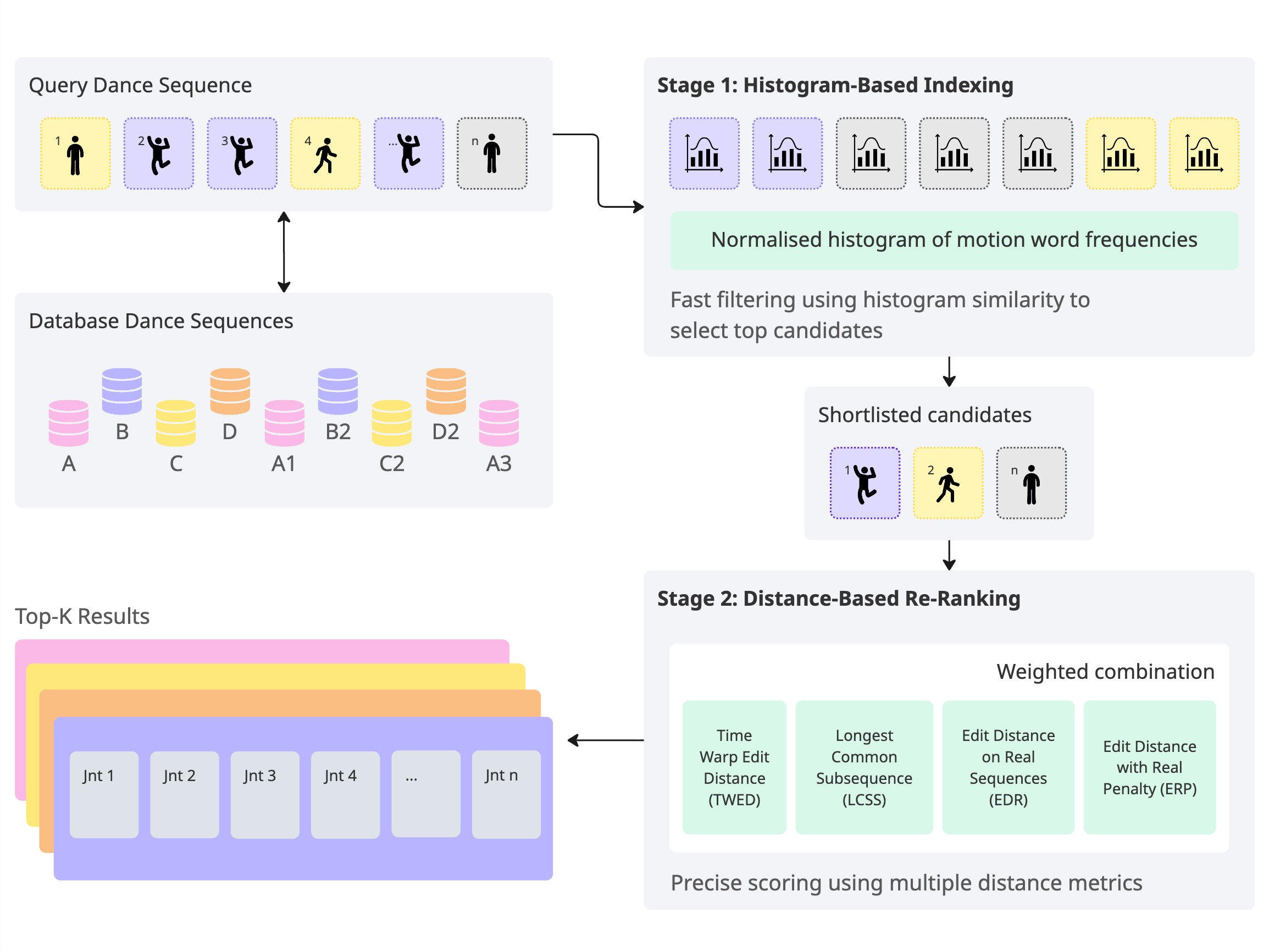}
\caption{Two-stage DRE architecture: histogram-based indexing (Stage~1) provides a shortlist for re-ranking via multi-metric temporal alignment (Stage~2). Complexity per stage: $O(NK)$ vs. $O(L M_q \bar{M})$.}
\label{fig:dre-arch}
\end{figure}

The \retrieval is a scalable retrieval framework for querying large collections of discrete motion word sequences extracted from dance videos. DRE employs a two-stage pipeline that decouples \emph{fast approximate search} from \emph{precise temporal alignment}, enabling real-time performance on thousands of dance sequences while maintaining near-optimal accuracy.

The design of DRE is guided by four interdependent principles that together ensure scalability, robustness, and extensibility of retrieval. \emph{(I) Statistical efficiency:} the histogram representation leverages sparsity in the motion word space to provide rapid, low-variance estimates of global similarity, maintaining high recall even for large databases. \emph{(II) Algorithmic complementarity:} the ensemble of edit-based temporal metrics compensates for the discretisation artefacts introduced by motion word quantisation, offering robustness to temporal misalignment and sampling irregularities. \emph{(III) Scalable abstraction:} by decoupling global (histogram) and local (alignment) similarity computations, DRE enables incremental index updates, cache reuse, and near-linear scalability with respect to database size. \emph{(IV) Composability:} the modular scoring pipeline admits seamless substitution of learned embeddings or neural similarity functions, preserving the architectural structure while allowing future optimisation through data-driven metric learning.

Given a query dance sequence $\mathbf{q} = [w_1, \ldots, w_M]$, where $w_i \in \{0, \ldots, K-1\}$ is a dance motion word from a vocabulary of size $K=512$, DRE retrieves the top-$k$ database sequences $\mathbf{s} \in \mathcal{D} = \{\mathbf{s}_1, \ldots, \mathbf{s}_N\}$ maximising a combined similarity function:
\begin{equation}
S(\mathbf{q}, \mathbf{s}) = \sum_m w_m \cdot \phi_m(\mathbf{q}, \mathbf{s}),
\end{equation}
where $\phi_m$ denotes a normalised similarity metric and $\sum_m w_m = 1$.  
This formulation unifies histogram-based retrieval and dynamic sequence alignment under a modular, extensible framework.

\subsection{Stage 1: Histogram-Based Indexing}
\textbf{Bag-of-Words Representation.} Each dance motion sequence $\mathbf{s} = [w_1, \ldots, w_M]$ is encoded as a normalised histogram of motion word frequencies:
\begin{equation}
\mathbf{h} \in \mathbb{R}^K, 
\qquad 
h_k = 
\frac{\sum_{i=1}^{M} \mathbb{1}[w_i = k]}
     {\left\|\sum_{k'} \mathbb{1}[w_i = k']\right\|_2},
\end{equation}
where $\mathbb{1}[\cdot]$ denotes the indicator function, and the histogram $\mathbf{h}$ is $\ell_2$ normalised to ensure scale invariance across sequences of varying length.
This representation captures the overall \emph{vocabulary} of motion primitives while discarding temporal ordering. As a result, visually or semantically similar dances yield comparable histograms even when executed with differing temporal dynamics or phase shifts. The invariance property makes this encoding suitable for coarse retrieval and clustering tasks where structural rather than temporal similarity is of primary interest.

\textbf{Cosine Similarity.}
Given normalized histograms $\mathbf{h}_q$ and $\mathbf{h}_s$, similarity between the query and a database sequence is measured using cosine similarity:
\begin{equation}
\text{sim}(\mathbf{q}, \mathbf{s}) = \mathbf{h}_q^\top \mathbf{h}_s.
\end{equation}
The index thus supports efficient candidate selection in $O(K)$ per comparison, invariant to sequence length.

\textbf{Adaptive Shortlisting.}
To balance recall and computational cost, DRE adaptively selects a shortlist size:
\begin{equation}
L = \max\left(\left\lfloor \frac{N}{2} \right\rfloor, \min(200, N)\right).
\end{equation}
This design guarantees that small databases are fully enumerated while large ones cap the shortlist to $L \leq 200$ candidates.

\textbf{Periodicity Detection.}
To capture the inherent rhythmicity of cyclic human motions such as walking or running, we integrate an \emph{optional periodicity detector} into the motion-word indexing pipeline. The detector estimates temporal regularity by computing the \emph{normalised autocorrelation function} over the temporal sequence of motion-word activations $\{w_t\}_{t=1}^{T}$:

\begin{equation}
\text{AC}(\tau) = 
\frac{\sum_{t=1}^{T-\tau} (w_t - \bar{w})(w_{t+\tau} - \bar{w})}
     {\sum_{t=1}^{T} (w_t - \bar{w})^2},
\end{equation}
where $\bar{w}$ denotes the mean activation and $\tau$ is the temporal lag.

A sequence is classified as \emph{periodic} if the autocorrelation exhibits multiple distinct local maxima above a confidence threshold $\theta = 0.6$:
\begin{align}
\text{isPeriodic}(\mathbf{s})
&=
\left|
\left\{
\tau :
\begin{array}{l}
\text{AC}(\tau) > \theta,\\[4pt]
\tau \text{ is local maximum}
\end{array}
\right\}
\right|
\geq 2.
\end{align}
This criterion discriminates genuine rhythmic behaviour from incidental repetitions or noise, thereby enabling the index to tag motion segments with a binary periodicity label. Although currently stored as auxiliary metadata, this signal can facilitate downstream tasks such as \emph{cyclic-motion retrieval}, \emph{gait normalisation}, or \emph{temporal alignment} across sequences. 

\subsection{Stage 2: Distance-Based Re-Ranking}
The shortlisted candidates are re-ranked using a weighted ensemble of robust sequence distance metrics, including \emph{Time Warp Edit Distance (TWED)}, \emph{Longest Common Subsequence (LCSS)}, \emph{Edit Distance on Real Sequences (EDR)}, and \emph{Edit Distance with Real Penalty (ERP)}. 


\textbf{Combined Scoring Function.} 
Given a query sequence $\mathbf{q}$ and a candidate $\mathbf{s}$ from the shortlist, we compute an aggregate similarity score as a convex combination of heterogeneous metrics capturing complementary structural properties:
\begin{equation}
\begin{aligned}
S(\mathbf{q}, \mathbf{s}) &= \sum_{m} w_m\, \phi_m(\mathbf{q}, \mathbf{s}), \\[4pt]
\text{s.t.} \quad
&\sum_m w_m = 1,\quad
w_m \ge 0.
\end{aligned}
\end{equation}
Here, $\phi_m$ denotes a normalised similarity under metric $m$, and $w_m$ its associated importance weight. The default configuration emphasises global appearance through the histogram term ($w_{\text{hist}}=0.3$) while allocating balanced capacity to alignment-based measures: TWED ($0.15$), LCSS ($0.15$), EDR ($0.15$), ERP ($0.10$), and a local-context $n$-gram term ($0.15$).  

By coupling global appearance cues with alignment-aware similarities, the formulation preserves semantic consistency even under non-uniform temporal dynamics that confound histogram-only approaches.

\textbf{Time Warp Edit Distance (TWED).}
The \emph{Time Warp Edit Distance (TWED)} extends the classical \emph{Dynamic Time Warping (DTW)} by explicitly modelling \emph{temporal elasticity} and \emph{irregular sampling}. Unlike DTW, which assumes uniformly sampled trajectories and penalises only alignment cost, TWED incorporates both temporal displacement
and edit operations, making it robust to real-world motion or event sequences with heterogeneous sampling rates.

Formally, given two temporal sequences $\mathbf{s}_1 = \{(s_1^i, t_1^i)\}_{i=1}^{T_1}$ and $\mathbf{s}_2 = \{(s_2^j, t_2^j)\}_{j=1}^{T_2}$, the TWED is defined recursively as:
\begin{equation}
\label{eq:twed}
\begin{aligned}
D(i,j) = \min
\begin{cases}
D(i\!-\!1,j\!-\!1)
 + d(s_1^i, s_2^j)
 + d(s_1^{i-1}, s_2^{j-1}) \\
\qquad +\, \nu |t_1^i - t_2^j|, \\[6pt]
D(i\!-\!1,j)
 + d(s_1^i, s_2^j)
 + \nu |t_1^i - t_2^j| \\
\qquad +\, \lambda, \\[6pt]
D(i,j\!-\!1)
 + d(s_1^i, s_2^j)
 + \nu |t_1^i - t_2^j| \\
\qquad +\, \lambda.
\end{cases}
\end{aligned}
\end{equation}
with boundary condition $D(0,0)=0$. Here, $d(a,b) = \mathbb{1}[a \neq b]$ denotes the discrete mismatch cost,
$\nu$ controls the \emph{temporal stiffness} (penalising misaligned timestamps), and $\lambda$ governs \emph{edit flexibility} by penalising insertion and deletion operations. 

To convert the TWED metric into a bounded similarity measure compatible with retrieval or kernel-based learning, we adopt an exponential transformation:
\begin{equation}
\begin{aligned}
\phi_{\text{TWED}}(\mathbf{q}, \mathbf{s})
&= \exp\!\left(
    -\frac{\text{TWED}(\mathbf{q}, \mathbf{s})}{2\bar{L}}
  \right), \\[4pt]
\bar{L} &= \tfrac{1}{2}\big(|\mathbf{q}| + |\mathbf{s}|\big).
\end{aligned}
\end{equation}
where $\bar{L}$ normalises by the average sequence length, ensuring scale invariance across variable-length trajectories.

Our implementation follows the original formulation with optimised dynamic programming and early termination heuristics for large-scale retrieval tasks

\textbf{Longest Common Subsequence (LCSS).}
The \emph{Longest Common Subsequence (LCSS)} measures sequence similarity by identifying the longest ordered subsequence of matching elements. Unlike DTW or TWED, which penalise all deviations, LCSS tolerates insertions, deletions, and local noise.

For sequences $\mathbf{s}_1 = \{s_1^i\}_{i=1}^{T_1}$ and $\mathbf{s}_2 = \{s_2^j\}_{j=1}^{T_2}$, the $\text{LCSS}(i,j)$ is defined recursively as:
\begin{equation}
\label{eq:lcss}
\mathcal{L}(i,j)=
\begin{cases}
0, & i=0 \text{ or } j=0, \\[4pt]
\mathcal{L}(i\!-\!1,j\!-\!1)+1, &
\substack{|s_1^i - s_2^j| \le \epsilon,\\ |i-j| \le \delta}, \\[6pt]
\max\{\mathcal{L}(i\!-\!1,j),\, \mathcal{L}(i,j\!-\!1)\}, &
\text{otherwise.}
\end{cases}
\end{equation}

We use $\epsilon = 0$ (exact match for discrete symbols) and $\delta = \infty$ (no temporal constraint). The normalised similarity is:
\begin{equation}
\phi_{\text{LCSS}}(\mathbf{q}, \mathbf{s})
= \frac{\text{LCSS}(|\mathbf{q}|,|\mathbf{s}|)}{\bar{L}},
\qquad
\bar{L} = \tfrac{1}{2}(|\mathbf{q}|+|\mathbf{s}|).
\end{equation}

\textbf{Edit Distance on Real Sequences (EDR).}
The \emph{Edit Distance on Real Sequences (EDR)} extends classical edit distance to continuous-valued data by introducing a matching threshold $\epsilon$ that tolerates small deviations between elements. It measures the minimum number of insertions, deletions, and substitutions required to transform one sequence into another under this tolerance.

Formally, for sequences $\mathbf{s}_1 = \{s_1^i\}_{i=1}^{T_1}$ and $\mathbf{s}_2 = \{s_2^j\}_{j=1}^{T_2}$:
\begin{equation}
\label{eq:edr}
\text{EDR}(i,j) =
\min \begin{cases}
\text{EDR}(i\!-\!1,j\!-\!1) + \text{cost}(s_1^i, s_2^j), \\[3pt]
\text{EDR}(i\!-\!1,j) + 1, \\[3pt]
\text{EDR}(i,j\!-\!1) + 1,
\end{cases}
\end{equation}
where $\text{cost}(a,b) = 0$ if $|a-b| \le \epsilon$, and $1$ otherwise. We use $\epsilon = 0$ for discrete representations.

The corresponding similarity is defined as:
\begin{equation}
\phi_{\text{EDR}}(\mathbf{q}, \mathbf{s})
= 1 - \frac{\text{EDR}(|\mathbf{q}|, |\mathbf{s}|)}
            {\max(|\mathbf{q}|, |\mathbf{s}|)}.
\end{equation}
EDR thus behaves as a normalised, noise-tolerant edit metric that aligns real-valued sequences while preserving interpretability in edit-distance terms.

\textbf{Edit distance with Real Penalty (ERP).}
The \emph{Edit Distance with Real Penalty (ERP)} extends classical edit distance by introducing a fixed \emph{gap value} and penalising deviations from it. This formulation maintains metric properties while modelling insertions and deletions in continuous-valued sequences.

For sequences $\mathbf{s}_1 = \{s_1^i\}_{i=1}^{T_1}$ and $\mathbf{s}_2 = \{s_2^j\}_{j=1}^{T_2}$, the recurrence is:
\begin{equation}
\label{eq:erp}
\text{ERP}(i,j) =
\min \begin{cases}
\text{ERP}(i\!-\!1,j\!-\!1) + d(s_1^i, s_2^j), \\[3pt]
\text{ERP}(i\!-\!1,j) + \beta |s_1^i - g|, \\[3pt]
\text{ERP}(i,j\!-\!1) + \beta |s_2^j - g|,
\end{cases}
\end{equation}
where $d(a,b)=|a-b|$ is the element-wise distance, $g$ is a constant reference (\emph{gap}) value, and $\beta$ controls the penalty strength. We use $g=0$ and $\beta=0.5$ in all experiments.

The similarity form is:
\begin{equation}
\phi_{\text{ERP}}(\mathbf{q}, \mathbf{s})
= \exp\!\left(
    -\frac{\text{ERP}(|\mathbf{q}|, |\mathbf{s}|)}{\bar{L}}
  \right),
\qquad
\bar{L} = \tfrac{1}{2}(|\mathbf{q}| + |\mathbf{s}|).
\end{equation}
ERP thus provides a smooth, gap-aware alignment measure suitable for continuous motion or feature trajectories.

\textbf{Dynamic Time Warping (DTW).}
The \emph{Dynamic Time Warping (DTW)} algorithm computes the optimal non-linear alignment between two temporal sequences by minimising the cumulative pairwise distance along a warping path. For discrete sequences
$\mathbf{s}_1 = \{s_1^i\}_{i=1}^{T_1}$ and
$\mathbf{s}_2 = \{s_2^j\}_{j=1}^{T_2}$, it is defined as:
\begin{equation}
\label{eq:dtw}
\text{DTW}(i,j) =
d(s_1^i, s_2^j) +
\min \begin{cases}
\text{DTW}(i\!-\!1,j\!-\!1), \\[3pt]
\text{DTW}(i\!-\!1,j), \\[3pt]
\text{DTW}(i,j\!-\!1),
\end{cases}
\end{equation}
where $d(a,b)=\mathbb{1}[a \neq b]$ denotes discrete mismatch cost.

Although DTW is widely used for temporal alignment, we do not employ it in final scoring due to its sensitivity to sequence length and lack of normalisation. Instead, we use length-normalised extensions such as TWED and ERP, which yield more consistent similarity scaling.

%% file: sec/5_experiments.tex
\section{Experiments }
\label{sec:experiments}

\subsection{\benchmark}
\label{sec:benchmark}

To complement AIST~\cite{aist-dance-db}, we introduce \benchmark, a large-scale benchmark of user-generated dance videos for motion retrieval under real-world conditions. 
The dataset is constructed by scraping \textit{TikTok} and \textit{YouTube Shorts} clips of \textit{Fortnite} in-game \textit{Emotes}, providing consistent choreography across varied character styles. 
Using metadata from a public \textit{Fortnite Emote List}, we automatically map video titles, tags, and descriptions to canonical emote labels.  

All clips are standardised to 30 fps and processed with CoMotion~\cite{newell2025comotion} to extract 3D skeletons, which are temporally smoothed and aligned. 

\benchmark bridges curated motion-capture datasets and noisy social-video corpora, enabling large-scale, label-aligned evaluation of dance fingerprinting and retrieval models.

\subsection{Experimental Setup}

We evaluate our motion word retrieval system using the AIST dataset~\cite{aist-dance-db}, which comprises ten dance genres (e.g., Break, Pop, House) performed by thirty dancers across sixty music tracks, yielding approximately 1,400 usable sequences after filtering for camera consistency. Each genre contains multiple choreographic instances that exhibit natural stylistic and temporal variations, thereby providing a robust testbed for retrieval consistency under intra-class diversity.

Following the methodology, we adopt a leave-one-out cross-validation protocol for the baseline experiments and a leave-$K$-out configuration for the enhanced variant. Unless otherwise stated, $K{=}1$ and the top-$3$ retrieved candidates are evaluated per query. All experiments employ precomputed motion word representations generated using the SMQ model, with inference performed on single-GPU hardware (NVIDIA H200). We compare two retrieval back-ends: (i) a histogram-indexed approximate retrieval method, and (ii) an exhaustive brute-force baseline.

\subsection{Evaluation Metrics}

Retrieval performance is quantified using three complementary metrics:
\begin{itemize}
    \item \emph{Mean Score}: the average of the rank-weighted score over all queries,
    \[
        \text{MeanScore} = \frac{1}{|\mathcal{Q}|}\sum_{i\in\mathcal{Q}} \text{score}(v_i),
    \]
    where the score decays with retrieval rank as $\{1.0, 0.5, 0.25, 0\}$ for ranks $1$–$3$ and beyond.
    \item \emph{Match Rate}: the proportion of queries whose top-$3$ retrievals contain at least one same-choreography performance.
    \item \emph{Rank Distribution}: the empirical distribution of best-match ranks across all queries.
\end{itemize}

\subsection{Baseline Results}

Table~\ref{tab:aist_results} reports results for both histogram-indexed and brute-force retrieval on the filtered AIST subset (ten genres, twenty samples per genre). The histogram index achieves $95.7\%$ of the brute-force performance at substantially reduced computational cost. Specifically, $65.3\%$ of queries retrieve a correct same-genre video within the top-$3$ candidates, and $39.0\%$ achieve perfect rank-$1$ matches.

\begin{table}[h]
\centering
\caption{Retrieval accuracy on AIST (ten genres, $\geq 2$ samples per class).}
\label{tab:aist_results}
\begin{tabular}{lccc}
\toprule
\textbf{Method} & \textbf{Mean} & \textbf{Match Rate} & \textbf{Rank1 (\%)} \\
\midrule
Histogram Index & 0.487 & 61.9 & 38.1 \\
Brute-Force     & 0.509 & 65.3 & 39.0 \\
\bottomrule
\end{tabular}
\end{table}

\subsection{Scalability to Larger Datasets}

We further evaluate on an extended dataset comprising $N{=}100$ dance categories with $M{=}50$ performance instances each, simulating real-world user-generated data (e.g., TikTok or YouTube). Using $K{=}10$ reference samples per class, the resulting query set of $4{,}000$ items tests the robustness of the learned motion vocabulary under high intra-class variance and noisy labels. We observe a mean normalised score $\overline{S}\geq0.5$, corresponding to rank-$1$ accuracy exceeding $40\%$, validating cross-dancer and cross-tempo generalisation.

%% file: sec/6_conclusion.tex
\section{Conclusion and Future Work}
\label{sec:conclusion}

We introduced \tool, an end-to-end framework for motion-based dance retrieval that transforms raw video into discrete, interpretable motion signatures. By combining Skeleton Motion Quantisation (SMQ) with Spatio-Temporal Transformers (STT), \tool constructs a structured motion vocabulary enabling efficient, large-scale retrieval through alignment-based metrics and a hybrid hierarchical index. Empirical results on \benchmark\ demonstrate strong cross-style generalisation and scalability, establishing a foundation for quantitative choreographic analysis.

Despite its effectiveness, the current framework remains limited by its loss of temporal ordering, lack of semantic grounding, and reliance on fixed metric weights. Future work will address these challenges by learning sequence-preserving embeddings that retain temporal dynamics, training neural similarity models to predict perceptual correspondence, and extending retrieval to hierarchical levels spanning motion phrases and full choreographies. Incorporating local temporal context (\eg, bigram or trigram motion patterns) further offers a path to richer temporal modelling without sacrificing interpretability.

Overall, \tool\ illustrates the potential of symbolic motion representations for bridging interpretability and scalability in dance analysis, paving the way toward learned, semantically structured motion retrieval.